\documentclass[a4paper]{article}
\RequirePackage{fix-cm}
\usepackage[english]{babel}
\usepackage[utf8x]{inputenc}
\usepackage[T1]{fontenc}

\usepackage[a4paper,top=3cm,bottom=2cm,left=3cm,right=3cm,marginparwidth=1.75cm]{geometry}

\usepackage{amsmath}
\usepackage{graphicx}
\usepackage[colorinlistoftodos]{todonotes}
\usepackage[colorlinks=true, allcolors=blue]{hyperref}
\usepackage{graphicx}
\usepackage{hyperref}

\usepackage{amsmath,array,graphicx}
\usepackage{kantlipsum}
\usepackage{booktabs,lipsum}

\usepackage{graphicx}
\usepackage[english]{babel}
\usepackage[utf8x]{inputenc}
\usepackage{amsthm}

\theoremstyle{definition}
\newtheorem{definition}{Definition}[section]

\usepackage{amsmath}

\usepackage{amsmath}
\usepackage{amssymb}
\usepackage{fixltx2e}
\usepackage{textcomp}
\usepackage{mathptmx}

\usepackage{booktabs}

\usepackage{flexisym}
\DeclareMathOperator*{\dprime}{\prime \prime}

\usepackage{mathtools}
\DeclarePairedDelimiter{\abs}{\lvert}{\rvert}

\usepackage{algorithm}
\usepackage[noend]{algpseudocode}

\usepackage{type1cm}  
\usepackage{stackengine}
\usepackage{mathptmx}     

\newcommand{\be}{\begin{definition}}
\newcommand{\ee}
{\end{definition}}

\title{Stochastic Dynamic 
Programming Heuristics for 
Influence Maximization-Revenue 
Optimization}

\author{Trisha Lawrence\\ 
Department of Mathematics and 
Statistics\\University of 
Saskatchewan \\
106 Wiggins Road\\ Saskatoon, SK S7N 5E6, CANADA}

\begin{document}
\maketitle

\begin{abstract}

The well-known Influence 
Maximization (IM) problem 
has been actively studied by 
researchers over the past 
decade, with emphasis on 
marketing and social 
networks. 
Existing 
research have obtained solutions 
to the IM problem 
by obtaining the influence 
spread 
and utilizing  the property of 
submodularity. This paper is 
based on a novel approach to the 
IM 
problem geared towards 
optimizing 
clicks and consequently revenue 
within  an
Online Social Network (OSN). 
Our approach 
diverts from existing 
approaches by adopting a novel, 
decision-making perspective 
through  
implementing 
Stochastic Dynamic Programming 
(SDP). 
Thus, 
we define a new problem 
Influence Maximization-Revenue 
Optimization (IM-RO) and 
propose SDP as a method in which 
this
problem can be solved. The SDP method has lucrative gains for an advertiser in terms of optimizing clicks and generating revenue however, one drawback to 
the method is its associated ``curse of 
dimensionality" particularly for 
problems 
involving a large state space. 
Thus, we introduce the Lawrence 
Degree 
Heuristic (LDH), Adaptive Hill-
Climbing (AHC) 
and Multistage Particle Swarm 
Optimization (MPSO) heuristics 
as methods which are orders of magnitude faster than the SDP method whilst achieving
near-optimal results. Through a 
comparative analysis on various synthetic and  real-world networks 
we 
present  the AHC and LDH as  
heuristics  well suited to to the IM-RO problem
in terms of
their accuracy, running times and 
scalability 
under ideal 
model parameters. In this 
paper we present a compelling 
survey on the SDP method as a 
practical and 
lucrative method for spreading 
information and optimizing revenue 
within the context of OSNs.
\end{abstract}

\section{Introduction}
Viral marketing 
possess lucrative 
advantages to advertising and 
marketing companies 
compared to traditional direct 
marketing strategies due to its 
ease of deployment and ability 
to effectively use customers 
themselves to encourage product 
preferences in others 
\cite{Meerman}. Viral 
marketing through online 
advertising   
accounts for a major source of 
revenue for many OSNs. For 
example, according to \cite{Facebook},
advertising continues to propel Facebook's revenue  
generation, accounting for $\$6.24$ billion, the majority 
source of income for Facebook in 
2016. OSNs utilize the advantage 
of viral marketing because one 
considers not only the
effect of marketing to a customer so that the customer purchases a product but also the customer's
influence in persuading other customers to purchase as well. \newline\indent
The focus of this paper
centers around the positioning of an advertiser's link delivered to a web page, that is to say, the 
placement of advertisement 
impressions. Advertising companies
have the task of placing 
impressions on pages to be 
displayed 
to its users. Thus, the
objective of the problem becomes to 
place impressions to OSN users in a 
way that maximizes the value to the 
advertiser. The problem known as 
the IM 
problem was first formally 
expressed in \cite{kempe}  as 
choosing a good initial set of 
nodes to target
in the context of influence models such as, the Independent 
Cascade, Linear Threshold and generalizations that followed ~\cite{ChenandCollins,GoyalL}. 
However, the problem of choosing an ideal 
set of customers in a network to market 
to in order to generate the maximum 
profit to the advertiser was first 
studied in ~\cite{Domingos,Rich}. Since its 
formal definition in \cite{kempe},  the 
IM problem has been actively
studied by researchers over the past 
decade and is not restricted to 
applications in marketing only but 
also
in healthcare, communication, 
education, 
agriculture, and 
epidemiology
~\cite{Kimura,Leskovec,Monroe,Singer}. For the
IM problem, OSN  users  are 
represented 
in a graph $G = (V,E)$, 
where the nodes
of $V$ represent the users and the 
edges 
in $E$ represent the 
relationships between
users. In \cite{kempe}, the 
problem was first defined as a 
discrete optimization 
problem and the term 
\emph{influence} of a set of
nodes $A$, denoted by 
$\sigma(A)$ was defined to be the 
expected
number of active nodes at the end of a diffusion process, 
given that $A$ was this initial active node set.
According to the work done in \cite{kempe}, the IM problem therefore
seeks to determine a parameter $K$, that is, to find a $K-
node$ set of maximum
influence; where $|A|\leq K$. It is an open question to 
compute this $K-node$ set and expected number of active nodes
$\sigma(A)$ by an efficient method, but 
very good 
estimates have been
proposed and obtained 
~\cite{WangEf,Chen,Leskovec}.
\newline\indent
This paper provides a novel 
approach to the IM problem and a 
formal definition 
to the model proposed in 
\cite{Lawrence}. We divert from 
all 
other existing approaches to IM 
and adopt a 
novel decision-making perspective 
primarily used 
in shortest paths and resource 
allocation 
problems 
\cite{BertsekasT,Levi,Nascimentop,Powell}. Thus, we
define a new problem, the IM-RO 
problem 
and implement SDP as the method 
in which this problem 
can be solved. The 
SDP 
method and its multistage 
attribute was 
demonstrated to generate 
lucrative gains to advertisers; 
causing over 
an  80\%  increase in the 
expected number of 
clicks
when evaluated on various networks. 
\newline\indent
Due to the complexity of the SDP method, we propose and analyze the 
LDH, AHC and MPSO algorithms as 
heuristics
employed to tackle the ``curse of 
dimensionality" associated with 
implementing SDP. Through conducting 
experiments on  synthetic networks, we 
demonstrate that all three methods 
achieved 
near-optimal solutions and are orders of 
magnitude faster than 
the SDP method. We provide a 
comparative 
analysis on various synthetic and  real-world networks and present 
the LDH and AHC as promising 
heuristics  in terms of
their  accuracy, running times and 
scalability under
suitable 
model parameters.
Although the MPSO heuristic 
generated 
the highest expected number of 
clicks when compared to the LDH 
and 
AHC heuristics,
it is unreliable and its
running time is too slow thus 
making it 
unfeasible for large graphs, i.e 
over 
500 nodes. 
Our results reveal the high 
potential 
of the LDH and AHC heuristics as 
an effective advertising strategy 
in providing near-optimal expected 
click values in minimal running 
times. \newline\indent 
Researchers have also sought to 
generate influence models which
capture the real-world influence 
outlined by the IM problem and to
determine node and edge
probabilities from real-graph data 
based on past-propagations 
~\cite{GoyalL,Goyal,Saito}. Effectively 
generating these 
influence models and 
computing their node and edge 
probabilities has also been an area 
widely researched. For the IM-RO problem 
we implement the Negative Influence 
model (NIM) and Graph Influence 
model (GIM) as the 
influence models which capture 
node and edge 
propagations among users within 
an OSN. At the end of each stage (specified time period) 
after users are placed with 
impressions, the node 
probabilities are updated based on the influence model used. The probability of a user clicking on an impression depends directly on their friend's behavior (whether their friend have clicked or not on an impression). 
\newline\indent
The paper is organized as 
follows. We briefly revisit the 
introduction of the SDP method 
to the IM problem in Section 
(\ref{sec:3}).
We then
introduce the proposed 
heuristics (LDH, AHC and MPSO 
algorithms)
in Section (\ref{sec:4}). In 
Section(\ref{sec:5}) we provide 
experimental 
results and a performance 
analysis for synthetic
networks  and real-world OSNs. 
We conclude the paper in Section 
{\ref{sec:6}}
summarizing the main 
contributions 
and directions for future work.

\section{Related Work}

The problem of 
selecting an ideal 
set of nodes in a graph or 
determining which set 
of 
users should be
marketed to in order to obtain the 
maximum expected profit from 
sales
was first studied in ~\cite{Domingos,Rich}. In these papers, 
the 
problem was 
viewed as trying to 
convince a subset of individuals 
to purchase a new product or 
innovation with the goal of 
generating further purchases over 
the 
entire network. In other words, the 
problem entailed
choosing specific users in a network 
which 
created a cascade over the entire 
network. Solutions to this problem 
comprised of both a 
non linear and a
linear probabilistic model that
optimized the revenue 
generated from sales. 
Subsequently, in \cite{kempe}
this optimization problem was 
defined as the 
IM problem 
and 
the emphasis shifted from maximizing 
the profit generated from sales to 
maximizing a cascade effect or 
the number of activated 
nodes at 
the end of a diffusion process in the 
context of diffusion 
models 
~\cite{Granovetter,Gomez,ChenandCollins,GoyalL}. 
Thresholds models have been 
studied in 
the context of sociological 
theory and social networks in 
~\cite{Granovetter,Macy,Jackson}. However, 
the generalization of the 
Linear Threshold  and Independent 
Cascade model  proposed 
in 
\cite{kempe} lies at 
the core of most threshold 
models for the 
IM problem .\newline\indent
Using the linear threshold and 
independent cascade models, the 
problem was shown to be NP hard 
in \cite{kempe}. Moreover, work 
done in \cite{Nemhauser} showed 
that using Linear 
Programming and the 
greedy algorithm, an approximate solution to the IM problem which was 
within $(1-1/e)$ 
of the optimal solution could be obtained.
$\newline\indent$
Approaches to solving 
influence maximization problems  have been put forward in 
~\cite{Narayanam,Galhotra}. 
Similar to the work done in 
~\cite{Domingos,Rich,Abassi}, we focus on the 
selection of an ideal set of nodes 
for the purpose of optimizing 
clicks and revenue to the advertiser. We 
divert from approaches to the 
problem that utilize influence 
spread and the theory of 
submodular functions as done in 
several 
papers
~\cite{KimApp,WangEf,Kimura,Liu,Leskovec,Chen,ChenandCollins,Bhagat,Cao,Galstyan,Hosseini,Wang,Wu} 
and focus on maximizing the 
expected gains for the 
advertiser. Our formulation to 
the problem formally known as the 
IM problem
is novel since in addition to 
adopting a decision-making 
perspective its main 
goal is to maximize the expected 
number or 
clicks or revenue to the advertiser. Thus we 
define a new 
problem, the  IM-RO 
problem and 
introduce SDP as 
the method in which this problem 
can be solved.

\begin{definition}[Influence 
Maximization - Revenue 
Optimization]
Given a  network modeled as 
graph $G=(V,E)$, a fixed number 
$M$ 
(impressions to be placed) and 
the probability $p_i$ of a user 
$i$ clicking on an impressions,
the problem seeks to find the 
optimal $M$ users to place 
impressions  so as to maximize 
the expected probability of 
clicking and ultimately revenue.

\end{definition}

\section{SDP for IM-RO}
\label{sec:3}
We consider the mulitstage SDP 
method to the IM, first 
introduced in \cite{Lawrence} 
and now defined as the IM-RO 
problem. 
The problem entails placing an 
integer amount, $m_{k}$, number 
of impressions to 
OSN users over $K$ stages, with
$k=0,1,2,...,K-1$ representing 
the 
number of stages to go. A user 
has 
two outcomes; to click or not 
click on an impression. The 
outcomes ${\cal V} \in \{0,1\}^N$ 
represent the set of 
possible outcome vectors  of a 
user clicking or not clicking on 
an impression with $|{\cal V}| = 
2^{m_k}$. The aim is to 
determine the number of 
impressions to be placed in each 
stage or impression-to-stage 
allocations $[m_{k-1}, m_{k-2}, m_{k-3},$
$...,m_{0}]$, 
and the optimal users $\vec{u*}$
that solves equation {\ref{optimal}}:

\begin{equation}\label{optimal}
\max_{\vec{u} \in \{0,1\}^N}  J_{k}= \sum_{k=0}^{K-1} J_{k-1}\prod_{i=1}^N u_k[i]\{p_k[i]v_{j}[i] + 
(1 - p_k[i])(1 - v_{j}[i])\} + 1 - u_k[i]
\end{equation}

\[ \mbox{subject to:}\;\;\; 
\sum_{i=1}^N u[i] = m_k \;\;\;\;\;\;\; \mbox{for the}  
j=1,...,2^{m_{k}} \mbox{outcomes}\]

where $\sum_{k=0}^{K-1} m_k = M$, 
is the total number of impressions 
allocated over $K$ stages such that 
$\vec{u}+\vec{x}_k \leq 
1$, as a user can only be given an 
impression once. 

$\{{u}_{k}[i] \in \{0,1\}\}$ and 
$\{{v}_{k}[i] \in \{0,1\}\}$ 
which represent user $i$ being not given 
or given and 
has not clicked or has  clicked on an 
impression respectively  (see 
\cite{Lawrence} for more 
details).

\begin{figure}[H]
\centering
    \includegraphics[width=70mm]
 {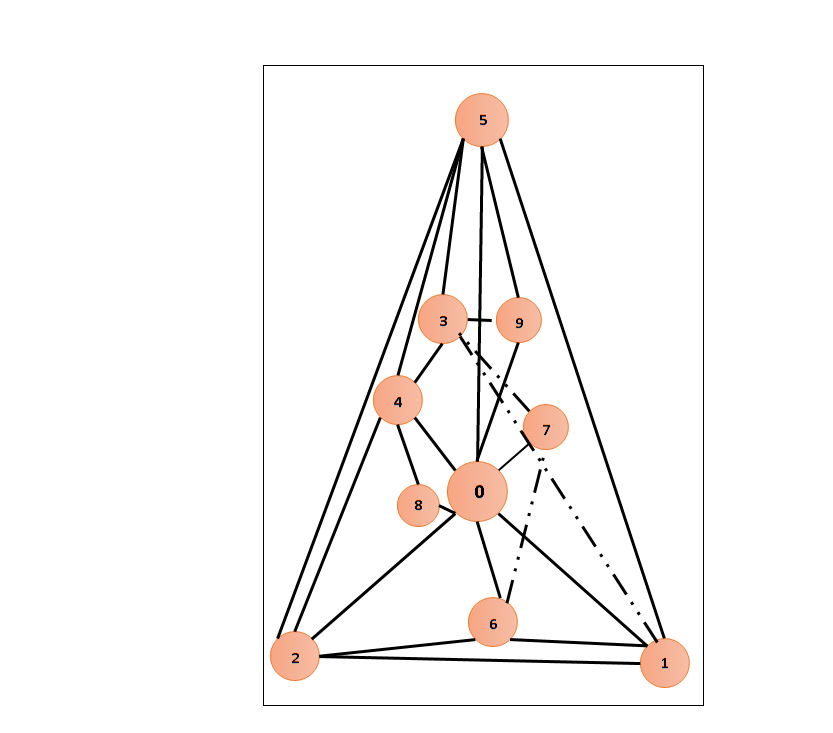}
\caption{A simple network on 10 nodes from graph generator}
\label{fig:1}      
\end{figure}

\begin{figure}[H]
\centering
    \includegraphics[width=70mm]
 {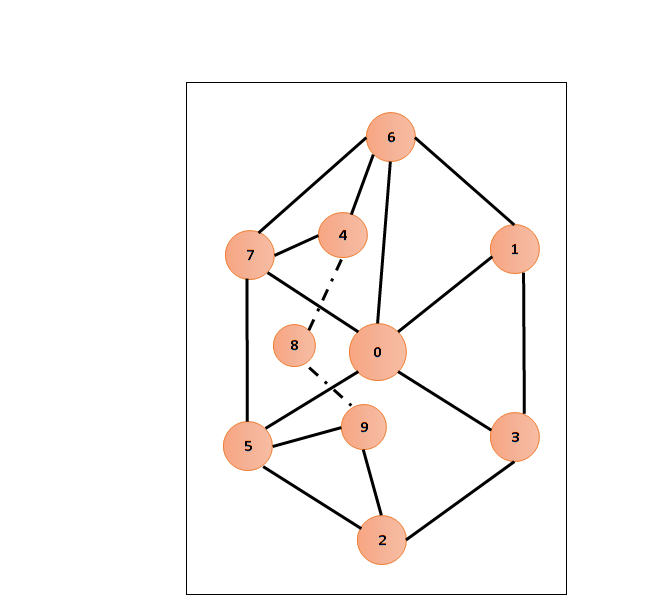}

\caption{A simple network on 10 nodes randomly drawn}
\label{fig:2}      
\end{figure}

We briefly revisit an evaluation 
of the SDP method through an 
analysis on two  simple 
networks, Figure 
(\ref{fig:1}) and Figure 
(\ref{fig:2}).
Tables 
(\ref{tab:one}, \ref{tab:two}  and
\ref{tab:three})
provide a 
concise survey of the SDP model on
these networks using both 
the GIM and NIM as the influence models by which probabilities are updated. The GIM is given by 
equation (\ref{graphinfluence}) and 
the NIM is given by equation (\ref{negative influence}).

\begin{equation} \label{graphinfluence}
p_k[i] = \max[0,\min[1, p_{0k} + (1-(1-\alpha \frac{y}{f})^{f})]]\;
\end{equation}

\begin{equation}\label{negative influence}
p_k[i] = \max[0, \min[1, p_{0k}+ \alpha \frac{y}{f} - \beta \frac{n}{f}]]
\end{equation}

For these models, $y$ 
represents 
the number of 
friends who have clicked on an impression,
$f$ 
represents the total number of friends of 
user $i$ and $p_{0k}$ 
represents a user's initial probability of 
clicking on an impression at the start of a $K$-stage
problem when $k=K-1$. This probability 
is a user's natural inclination 
for clicking on an impression in the absence 
of 
any influence from friends.
$\alpha$ and $\beta$ are influence constants 
where
$\beta$ is the negative influence 
constant associated with $n$, the number of 
users who 
have been given impressions and have 
not clicked on them.

\begin{table}[H]
\label{tab:one} 
\caption{6 impressions varying stages on Figure \ref{fig:1}}
\label{tab:one}    
\begin{tabular*}{\linewidth}{@{\extracolsep{\fill}}p{0.05\linewidth}p{0.05\linewidth}p{0.05\linewidth}p{0.05\linewidth}p{0.1\linewidth}p{0.1\linewidth}@{}}

\toprule
Influence Model &Stages& Allocation  & User & Expected Clicks & Time \\

\midrule

GIM & 2 & [2,4] & 0,4 &2.12 & 300\\
&3 & [1,2,3] & 1 & 2.36 & 1240\\
&4 & [1,1,2,2] & 0 & 2.56 & 2,720 \\
&5 & [1,1,1,2,1] & 8 & 2.64 & 6,830 \\
&6 & [1,1,1,1,1,1] & 1 & 2.69 & 12,540 \\
NIM & 2 & [1,2] & 0 &0.96 & 244\\
&3 & [2,1,3] & 1,2 & 1.53 & 1,360\\
&4 & [2,1,3,0] & 1,2 & 1.53 & 2,320 \\
&5 & [2,1,3,0,0] & 1,2 & 1.53 & 6,560 \\
&6 & [2,1,3,0,0,0]  & 1,2 & 1.53 &  12760\\
\bottomrule
\end{tabular*}

\end{table}

\begin{table}[H]
\caption{6 impressions varying stages on Figure \ref{fig:2}}
\label{tab:two}        
\begin{tabular*}{\linewidth}{@{\extracolsep{\fill}}p{0.05\linewidth}p{0.05\linewidth}p{0.05\linewidth}p{0.05\linewidth}p{0.1\linewidth}p{0.1\linewidth}@{}}

\toprule
Influence Model &Stages& Allocation  & User & Expected Clicks & Time \\

\midrule

GIM & 2 & [3,3] & 0,2,4 &1.94 & 240\\
&3 & [2,2,2] & 0,4 & 2.08 & 1,300\\
&4 & [1,2,2,1] & 1 & 2.13 & 3,540 \\
&5 & [1,1,2,1,1] & 1 & 2.19 & 5,880 \\
&6 & [1,1,1,1,1,1] & 3 & 2.22 & 12,960 \\
NIM & 2 & [2,4] & 2,3 & 0.96 & 230\\
&3 & [2,3,1] & 2,3 & 1.58 & 1,310\\
&4 & [1,1,3,1] & 2 & 1.58 & 3,670 \\
&5 & [1,1,3,1,0] & 2 & 1.58 & 6,110 \\
&6 & [1,1,3,1,0,0]  & 2 & 1.58 &  12,930\\
\bottomrule
\end{tabular*}

\end{table}

\begin{table}[H]
\caption{Increasing the number 
of impressions in 3 stages on Figure \ref{fig:1}}
\label{tab:three}       
\begin{tabular*}{\linewidth}{@{\extracolsep{\fill}}p{0.05\linewidth}p{0.05\linewidth}p{0.05\linewidth}p{0.05\linewidth}p{0.1\linewidth}p{0.1\linewidth}@{}}

\toprule
Influence Model &Stages& Allocation  & User & Expected Clicks & Time \\

\midrule
GIM &3 & [1,1,1] & 0 &1.04 & 18\\
&4 & [1,1,2] & 0 & 1.48 & 90 \\
&5 & [1,1,3] & 0 & 1.91 & 440\\
&6 & [1,2,3]  & 0 & 2.36 &  1331\\
NIM &3 & [1,1,1] & 3 &0.8 & 14\\
&4 & [1,2,1] & 3 & 1.06 &110  \\
&5 & [1,3,1] & 0 & 1.3 & 473 \\
&6 & [2,4]  & 7,8 & 1.53 &  1300\\
\bottomrule
\end{tabular*}

\end{table}

For a 
single stage problem involving 6 
impressions with $p_{0k}=0.25$, the 
optimal expected 
number of clicks is calculated as 
1.5.  The optimal 
expected number of clicks 
determined 
by the SDP method with $\alpha =0.25$, 
$\beta=0.25$ and under the GIM is 2.69, that is 
an increase 
of approximately 80\% percent.
These results have considerable 
gains for 
the advertiser in terms of 
spreading information and 
optimizing 
revenue. We note 
that for the NIM, the optimal expected 
number of clicks is achieved at 3 stages 
in both networks
whilst for the GIM the optimal expected click 
value increases as the number of stages 
increases. 
However, the running times to achieve 
this results is unfeasible,  
especially 
for large networks hence the need for 
computationally less extensive 
heuristic solutions which 
achieve near 
optimal results. For both the 
GIM 
and NIM model a significant 
increase 
in the optimal expected number 
of 
clicks can be achieved at 3 
stages in 
reasonable time. Another 
interesting 
fact, is the drastic 
increase in running times caused 
by adding a single 
impression. When $M=5$ and 
$K=3$, the SDP method achieves 
the optimal solution in 
approximately 7 minutes on these 
simple 
networks. \newline\indent For an asymptotic analysis 
on the SDP method, 
we consider a 2-stage problem with 
$M$ 
impressions to be placed to its 
users. If we consider the  
impression-to-stage allocation [1, M-1], then  there are
$\binom{N}{1}$ possible combinations 
of 
users to choose from for this. For 
[2, M-2], there are  $\binom{N}{2}$ 
possible combinations 
of 
users to choose from and  $\binom{N}
{3}$  possible combinations of users 
to choose from for [3, M-3]. If we 
continue counting the steps in this 
manner until the last impression-to- 
stage allocation  $[M, 0]$, then 
using the  Binomial Theorem, we can 
prove that
that $2^N$ is an upper bound on the 
number of steps to attain the 
optimal solution. Hence the SDP 
method 
has a complexity of  $\mathcal{O}
(2^n)$ in its worst case. For large 
graphs, this proves to be 
intractable.
In order to reduce its complexity 
and evaluate 
the performance of the SDP 
method on larger 
networks, we propose heuristics 
which leverage 
on the optimality of the SDP 
method whilst reducing its 
complexity.\newline\indent
Below we describe three heuristics, the LDH, 
AHC and MPSO that
adopt the multistage aspect of 
the SDP method. The MPSO, however
is the least reliable in terms 
of its accuracy 
compared to the LDH and AHC   
since its state space comprises of 
all 
possible predetermined users in each 
impression to stage 
allocation and their associated expected number 
of clicks. This is an essential 
characteristic of the SDP method, and LDH and AHC heuristics in 
attaining the optimal and near-optimal solution.

\section{Heuristics }\label{sec:4}

\subsection{LDH}

We begin by introducing the LDH as a method which reduces 
the complexity of the SDP method by reducing its branching 
factor. For a given 2 or 3 stage 
problem, the LDH generates the impression-to-stage 
allocation [1, M-1] or [1, 1, M-2] respectively.
Next, the optimal expected number of clicks is computed 
for this impression-to-stage allocation using equation 
$(\ref{optimal})$ and with users $(\vec{u}^L)$.
Here $\vec{u}^L$ is the optimal 
solution to 
the IM-RO 
problem in which $w^*$, the node 
of the 
highest valency 
in the graph is selected at the first stage 
when $k=K-1$. 
The 
inspiration for the LDH is based on the 
efficiency of 
well 
known high degree heuristics  in 
\cite{kempe} as well as the 
experimental findings 
of Section(\ref{sec:3}) in which 
the 
optimal solution 
was achieved in 3 stages. As the 
LDH expands 
only one 
node corresponding to either [1, M-1] or 
[1, 1, M-2], its 
complexity is $\mathcal{O}(1)$, 
which is a 
drastic 
reduction to the complexity of 
the SDP method.

\begin{algorithm}
\caption{LDH}\label{LDH}
\begin{algorithmic}[1]
\Procedure {LDH; Input : G=(V,E), 
number of impressions, $M$, 
number of 
stages, $K$, number of iterations, $n$ }{}
\If {$K=2$ }
\Return 
\State impression-to-stage allocation [1, M-1] 
\Else 

\State impression-to-stage-allocation [1, 1, M-2].
\EndIf

 \State select the first node for this impression-to-stage allocation as the node with the highest degree, $w*$, compute the expected number of clicks generated.
 
\State \textbf{return} solution

\EndProcedure
\end{algorithmic}
\end{algorithm}

\subsection{AHC}

The hill-climbing search algorithm often referred 
to as the 
greedy hill-climbing algorithm is an example 
of a local search algorithm that operates by 
expanding a single node and navigating to 
neighboring nodes 
with the goal of finding the global 
minimum/maximum, if 
one exists. The general hill-
climbing algorithm and its variants have been proposed in ~\cite{Tsang,Davis}. Moreover, for the IM problem the greedy 
hill-climbing  
algorithm and improvements of this algorithm have 
been proposed  in several papers
~\cite{kempe,Kimura,Leskovec,Liu,WangEf}.
\newline\indent
We implement an adaptive hill-
climbing technique to 
the IM-RO problem with the 
functionality of the 
general hill-climbing algorithm, 
however the 
algorithm expands nodes 
corresponding to the 
impression-to-stage allocation 
[1,M-1] or [1, M-2, 1] for a 
given 2 or 3 stage problem 
respectively. The 
first node to expand in the $(K-1)$th stage-to-go 
is chosen randomly. Based on the 
click 
outcomes, the 
probabilities over the 
entire network are updated using 
either the NIM or 
GIM and the expected 
number of clicks computed as in 
the SDP method. 
For the AHC algorithm, each time 
a node is 
randomly chosen in the $(K-1) th$ stage to go and the 
expected number of clicks 
computed for the 
allocation using equation 
(\ref{optimal}), its value is 
compared to the previous 
value computed. The AHC  
algorithm
continues randomly expanding 
nodes in the $(K-1)th$  stage-
to-go and computing their 
associated optimal expected 
number of clicks for a specified 
number of iterations $n$.
In general, the hill-climbing 
algorithm does not guarantee the 
optimal solution, however has an 
$O(1)$ memory and is  quite 
efficient. We provide the hill 
climbing algorithm adapted to 
the IM-RO problem as follows:

\begin{algorithm}
\caption{AHC}\label{hillclimb}
\begin{algorithmic}[1]
\Procedure {AHC; Input : G=(V,E), 
number of impressions,$M$, number of 
stages, $K$, number of iterations, $n$ }{}
\If {$K=2$ }
\Return 
\State impression-to-stage allocation [1, M-1] 
\Else 

\State impression-to-stage-allocation [1, M-2,1].
\EndIf
 \For{ iteration $t = 1...n$ } 
 \State select randomly the first user for this impression-to-stage allocation and obtain the expected number of clicks generated.
 \If {current solution $\geq$ previous solution }
\Return 
\State current solution
\Else 

\State \textbf{return} previous solution
\EndIf
 
      \EndFor
\EndProcedure
\end{algorithmic}
\end{algorithm}

\subsection{MPSO}

Particle Swarm Optimization 
(PSO), was first 
proposed
as one of 
the swarm intelligence 
algorithms for optimizing 
continuous nonlinear functions in 
\cite{Eberhart}. PSO is an 
algorithm that is modeled on the 
social behavior of swarming 
observed in insects, fishes and 
birds \cite{Swarm Intelligence}. 
The main idea of PSO originated 
from the movement of 
bird flocks, in which the 
algorithm can find the 
optimal 
solution in the search space 
just like a flock of bird 
searching for its food. For the original continuous space
PSO algorithm proposed 
in \cite{Eberhart}, the particles 
cooperated with each other in a 
global optimum and $n$-dimensional 
search space in order to move to better positions.. The position vector $X_{i}$ is 
used to  denote the 
current solution of 
particle $i$ whilst the velocity vector $V_{i}$ 
is 
used to provide the direction of the $i-th$ 
particle and 
adjust the particle’s position to the 
optimal solution. Various researchers have extended the 
original PSO algorithm proposed in 
\cite{Eberhart} to discrete 
optimization problems 
~\cite{Clerc,Wang,Salman,Sha}. The first of 
this kind was 
the binary particle swarm (BPSO) 
proposed in 
\cite{binary}. Similar to the 
continuous space 
PSO algorithm, the discrete 
space PSO algorithm 
involves the following 
probability update 
rules:

\begin{equation}\label{eq:updatepos}
X_{i}^{j+1} = X_{i}^{j} +V_{i}^{j+1}
\end{equation}

\begin{equation}\label{eq:updatevel}
V_{i}^{j+1}= v_{i}^{j} + c_{1}r_{1}^{j}( PBest_{i}^{j}-X_{i}^j) + c_2 r_{1}^{j}(GBest^j-X_{i}^{j})
\end{equation}

The $i-th$ 
particle maintains both a 
position and velocity over $n$ iterations given by 
$X_{i}(x_{i}^1, 
x_{i}^2,...,x_{i}^n)$ and $V_{i}
(v_{i}^1, v_{i}^2,...,v_{i}^n)$ respectively, 
where $PBest_{i}$ 
$(pbest_{i}^{1},$ $pbest_{i}^{2},...,pbest_{i}^{n}
)$ is the vector representing the personal best 
solution of the $i-th$ particle and $GBest_{i}$ 
$(gbest^{1}, gbest^{2},...,gbest^{n} )$, the 
global best solution obtained by the entire 
swarm. $c_1$ and $c_2$ are parameters which 
weigh each particles own experience and the the 
entire swarm respectively whilst, $r_1$, $r_2$ are constants such that, $r_1$, $r_2$ 
$\in$ [0,1]. At each iteration, the
particle's velocity is updated 
by using its own 
search experience and the 
experience of the 
entire swarm as it flies to a 
new search 
position.\newline\indent
For the implementation of the 
MPSO algorithm, 
the state space comprises of the 
set of 
possible predetermined users in 
each impression to stage 
allocation
with their corresponding 
expected number of clicks. We 
modify and make use of 
a key concept called a Swap 
Operator proposed in \cite{Wang} 
to handle discrete type PSO 
problems. 
For the implementation of the 
MPSO algorithm, a  solution 
set  $S$ can be described as a  
specific impression to stage 
allocation in which 
all of the users are identified.
We define a Swap operator $SO (j, 
i)$ as intechanging user i with the user in the $j$-th position, $SO ^{+k} (j , 
i)$ as addding user $i$ to the the $j$ th position in the $k-th$ stage to go and  $SO ^{-k} (j , i)$  as removing  user $i$ to and from the 
$j-th$ position in the $k-th$ stage to go.
Using the these swap operators we can redefine 
addition on the
solution sets $S$ with 
a new solution $ S'$. That is,

\begin{equation}\label{swap0}
S' = S + SO(j , 
i)
\end{equation}

\begin{equation}\label{swap1}
S' = S + SO ^{+k} (j , 
i)
\end{equation}

\begin{equation}\label{swap2}
S' = S + SO ^{-k} (j , i)
\end{equation}

A swap sequence $SS$, is a 
sequence made up 
of one or more of the following 
Swap Operators as 
defined in equations 
(\ref{swap0},
({\ref{swap1}} and 
\ref{swap2}).
We redefine subtraction,  $S_{1}- 
S_{2}$ on two solutions $S_{1}$ 
and $S_{2}$ as the Swap Sequence 
$SS$ 
acting on the solution $S_{2}$ in order to 
obtain solution $S_{1}$. \newline\indent
For example,
consider a SDP formulation 
of the IM-RO problem involving 4 impressions 
and 2 stages, with two solutions $S_{1}$ and $S_{2}$:
%
$S_{1}$=[2, 2] with users 1,2  
in the first stage and 3,5 in the second stage.

$S_{2}$ = 
[1,3] with users 5 in the first stages and 2,3,1 in the 
second stage.

We can apply the Swap Operator $SO^{-0}(1, 2)$ to 
$S_{2}$  
removing user 2 from the first position  to 
obtain a new solution 
$S_{2} 
^{\textprime}$ = [1,2] with user 5 in the first 
stage and use 3,1 in the second stage. The 
second Swap Operator $SO^{+1}(2,2)$ can be applied to 
$S_{2} ^{\textprime}$ where user 2 is added to position 
2 in order to obtain a new 
solution $S_{2}^{\dprime}$=$[2,2]$ with user 5,2 in the 
first stage and users 3, 1 in the second stage. The 
third swap  operator $SO(1, 1)$ is applied to 
$S_{2}^{\dprime}$ and interchanges the user 
in position 1 with user 1. Thus $S_{2}^{\prime \prime \prime}$ = [2,2] with user 1,2 in the first stage and 
3,5 in the second stage. Hence, a swap sequence
$SS$ with the least number of 
operators for $S_{1}-S_{2}$  is $SS$= $SO^{-0}(1, 2),$ 
$SO^{+1}(2, 2)$ $,SO(1,1)$. In 
implementing the MPSO, the 
velocity is updated using 
equation $(\ref{eq:updatevel})$ 
and applying the relevant swap sequences $SS$. We 
provide an 
algorithm, Algorithm $(\ref{MPSO}) $ for the procedure as follows:

\begin{algorithm}
\caption{MPSO}\label{MPSO}
\begin{algorithmic}[1]
\Procedure {MPSO; Input: G= (V, E), 
swarm size, $\abs{n_{s}}$,  number of 
iterations $n$, number of stages $K$, $r_{1} c_{1}, 
r_{2}c_{2}$  }{}
\For {particle $i =1 $ to $\abs{n_{s}}$ }
\State initialize position $\vec{X}$ $\leftarrow$ $\vec{u}_{k}$ $\in n_{s}$
\State initialize $\textbf{PBest}$ $\vec{X}$ $\leftarrow$ $\vec{u}_{k}$ $\in n_{s}$
\State initialize velocity $\vec{V}$ $\leftarrow$ 0
 \EndFor
 \State Based on click values, select the global best $\textbf{GBest}$
 \For{ iteration $t = 1...n$ } 
 \State Update velocity $\vec{V}$.
 \State Update position $\vec{X}$.
  \State Update $\textbf{PBest}$  and select $\textbf{GBest}$ in this iteration. 
 \State Update $\textbf{GBest}$ as the best position found so far.
      \EndFor
\State \textbf{return}  $\textbf{Gbest}$ and $\vec{u}_{k}$ as the best position (solution) to the IM-RO problem.
\EndProcedure
\end{algorithmic}
\end{algorithm}

\section{Experiments}\label{sec:5}

We evaluated the effectiveness 
of the proposed heuristics using 
synthetic and real-world OSNs.

\subsection{Datasets}

We employed various synthetic 
networks and two 
real-world OSNs represented 
as graphs to analyze each 
method. Synthetic networks of 
various sizes 10, 50,100, 500, 
1000, 2000, 4000,  4500 and 5200 
were generated 
using  a pseudo random 
number generator as done in 
\cite{Matsumoto}. From a sample 
of 10 generated synthetic graphs, the 
average node degree was found to 
be  at least 60 \% of the number 
of nodes in the 
graphs.\newline\indent
In addition to these networks, we 
utilized two  real-
world OSNs
Flickr and Epinions obtained 
from the Social Computing Data 
Repository in \cite{Zafarani} and 
the 
Stanford Network 
Analysis Platform  in
\cite{LeskovecS} respectively.
The OSN, Flickr is
is an  image hosting and video 
sharing website where users can 
share images among each other. 
In this network 
"1,2" is used to represent the 
friendship relationship between  
the
user id 1 and the user id 2. The 
entire dataset consists of  
80,513 nodes, from this 
we extracted two datasets, FL1 
comprising of  11,098 nodes and 
FL2, comprising of 20,217 nodes 
each with an average 
node degree of 2 nodes for the 
purpose of 
evaluating each 
heuristics.\newline\indent
Epinions is a customer review  OSN  in which users rate 
various products that are 
purchased on Ebay. The entire 
dataset 
consists of  75,879 nodes, from which, we extracted a dataset of 
4,382 nodes  with 
an average node degree of 3 
nodes and refer to this 
dataset as Ep.

\subsection{Experimental 
Settings}

Influence models for the IM problem 
can be 
described as models which 
capture real-world propagations 
or the spread of information 
among users within a 
network. In addition to the 
diffusion models; the Linear 
Threshold and 
Independent Cascade models 
defined in \cite{kempe}, 
influence models that determine 
node and edge 
probabilities have been 
proposed in 
~\cite{Domingos,Rich,Galhotra,GoyalL,Cao,Chakrabarti}.
For the IM-RO problem we 
introduce the GIM equation 
(\ref{graphinfluence}) and NIM 
(\ref{negative influence}) as 
the pertinent 
influence models by which probabilities are updated at the end of each stage. The SDP method 
for the problem adopts a 
multistage approach and at each 
stage 
users are provided 
with advertising links or 
impressions. At the end of each 
stage, the outcomes or whether a 
user has clicked or not are
determined and this information 
is utilized in the influence 
models to update the 
probabilities for future 
stages. The objective thus becomes to 
determine the number of 
impressions to be at placed at 
each stage and the users to 
place impressions to, so as to 
maximize the number of 
purchases. A user clicking on an 
impression is equated to a user 
purchasing a product, therefore 
optimizing the revenue generated 
is identical to optimizing the 
expected number of clicks. 
A user's initial probability of 
clicking, $p_{0}$ was 
arbitrarily set to be 
$0.25$ for these experiments, 
$\alpha$ and $\beta$ were also 
arbitrarily 
set to be 0.25.
However for future work, we will 
demonstrate that 
$p_{0}$
can be effectively estimated 
using data mining techniques. \newline\indent
All our experimentation was 
undertaken on a server with 
8GB of RAM and i3 Processor. The 
SDP method, LDH, AHC and MPSO heuristics were implemented from 
scratch using a Python version 
2.7 (64 bit)  with an average of 
10 runs 
taken for each experiment.

\subsection{Performance Analysis 
on Synthetic Networks}

\begin{table}[H]
\caption{Results for 5 impressions and 3 stages with AHC}
\label{hilliterations}       
\begin{tabular*}{\linewidth}{@{\extracolsep{\fill}}p{0.05\linewidth}p{0.05\linewidth}p{0.05\linewidth}p{0.05\linewidth}p{0.1\linewidth}@{}}

\toprule
Influence Model &Graph Size& Iteration (n)  & Optimal Clicks & Time (secs)  \\

\midrule
GIM&50 &1  &1.675 &5 \\
 &&5 & 1.688 & 10 \\
 &&10 & 1.727& 22 \\
 & &20 & 1.718& 38 \\
 & &50  & 1.821& 78\\
&500 & 1 &1.673 & 17 \\
& & 5 & 1.673 & 98 \\
& & 10 & 1.673 & 280 \\
& & 20  & 1.673 & 360\\
& & 50 & 1.709 & 981 \\
& & 100 & 1.721& 2,348 \\
& 2000 & 1 & 1.673 & 231 \\
& &5& 1.673 & 1,237 \\
& &10 & 1.673 & 2,257\\
&& 20 & 1.71 & 3,615 \\
& 4500 &1 & 1.672 & 1,218\\
& & 5 & 1.672 & 7215 \\
& & 10 & 1.672 & 14,427 \\
NIM&50 &1  & 1.260 & 5\\
 &&5 & 1.261 & 8\\
 & &10 & 1.262& 15\\
 & &20 & 1.262& 25\\
 & &50  & 1.263& 53\\
&500 & 1&  1.251  & 23 \\
& & 5 & 1.251& 85\\
& & 10 & 1.251& 187\\
& & 20  & 1.251 &327 \\
& &50& 1.251 & 877\\
& & 100 & 1.252& 2,060 \\
& 2000 & 1 & 1.250 & 215\\
& &5& 1.250 & 1,311 \\
& &10 & 1.250 &2,276 \\
&& 20 & 1.250 &3,616 \\
&4500 &1 & 1.250 & 1,198 \\
& & 5 & 1.250 &7,228 \\
& & 10 & 1.250 & 18,031 \\

\bottomrule
\end{tabular*}

\end{table}

\begin{table}[H]
\centering
\caption{3 stages, with LDH under GIM }
\label{Law}      
\begin{tabular}{c*{5}{>{$}c<{$}}}

\toprule

Datasets & Impressions  & Expected Clicks &  Time (secs)  \\

\midrule
500 & 5 & 1.45 & 19\\
1000 & 5& 1.45 & 58 \\
2000 & 5 & 1.45 & 230 \\
4500 & 5 & 1.45 & 1874\\
5200 & 5 &1.45  &  3,489\\

\bottomrule
\end{tabular}

\end{table}

\begin{table}[H]
\caption{Results for 5 impressions in 2 for MPSO under NIM}
\label{psocurrent}     
\begin{tabular*}{\linewidth}{@{\extracolsep{\fill}}p{0.05\linewidth}p{0.05\linewidth}p{0.05\linewidth}p{0.05\linewidth}p{0.1\linewidth}@{}}

\toprule
Graph Size & Swarm Size  & Iteration  & Optimal Clicks & Time (secs)  \\

\midrule
50 &10 &1  & 1.260 & 5\\
 && 10 & 1.261 & 50\\
 & & 20 & 1.261& 63\\
 & &40 & 1.261& 117\\
 & &80  & 1.261& 261\\
  & &100  & 1.261& 275\\
&50 & 1&  1.262& 417  \\
& & 10 & 1.262& 154\\
& & 20 & 1.262& 294\\
& & 40  & 1.262 & 535 \\
& &80& 1.262 & 1335\\
& 100 & 1 & 1.262 & 63\\
& &10& 1.262 &284 \\
& &20 & 1.262 &608 \\
&& 40 & 1.262 &1,062 \\
&& 80 & 1.262 &2,176 \\
500&10 &1  & 1.251 &114 \\
 &&10 & 1.251
 & 357 \\
 &&20 & 1.251& 697 \\
 && 40 & 1.251 & 1,706\\
 & &80 & 1.251& 3,606 \\
&20 & 1 & 1.251 & 179 \\
& & 10 & 1.251 & 973 \\
& & 20 & 1.251& 2,221 \\
& & 40  & 1.251 & 3,617\\
& & 80 & 1.251
& 7,440 \\
& 50 & 1 & 1.251 & 467 \\
& &10& 1.251 & 2,258 \\
& &20 &1.251 & 4,920\\
&& 40 & 1.251
 & 7,257 \\
2000&10 &1  &  1.250
&1,998 \\ 
& &10& 1.250
 &10,838  \\

\bottomrule
\end{tabular*}

\end{table}

The results indicated in Table 
(\ref{Law}) convey the 
optimal expected number of clicks and 
running times of the LDH under the 
GIM. As shown in Table 
(\ref{Law}), the LDH is orders of 
magnitude faster than the SDP  
method achieving a ``good"  
solution of 1.45 in less than an hour 
on a synthetic graph of  5200 nodes. For now, we can think of 
a ``good" 
solution as a solution that is at 
least as high as the 
value obtained by placing all 
the impressions in one stage, 
however, for future work we will 
obtain an upper 
bound on the optimal solution, as 
this will provide greater insights 
into reasonable solutions and
how well these heuristics perform on 
large graphs. We note that the 
optimal expected 
number of clicks determined by the 
SDP method on a graph of 10 nodes 
was found to be 1.91 under identical 
model parameters of this experiment. We further note 
that an 
increase in the value of 
$\alpha$, even when $p_{0}$ 
is assigned small values $(\leq 0.5)$, results in  
an 
increase in the expected number 
of clicks. Hence we propose the 
LDH method as a reasonable and 
promising method which leverages on the accuracy of the 
SDP method whilst reducing its 
complexity. \newline\indent
To evaluate 
the performance of the AHC 
algorithm, we 
varied
the graph 
sizes and number of iterations. 
The results in Table 
(\ref{hilliterations}) indicate 
that the optimal 
expected number of clicks increases 
with the number of iterations, 
particularly for large values of 
$n$, 
that is, $n\geq 50$ as seen in the 
graphs of 
50 and 500 nodes. The AHC 
algorithm generates a value of 
1.250 for a 
graph of 2000 nodes in less than 50 
iterations and 1.252 
for a 
graph for a graph of 500 nodes in 
100 iterations, both values 
in less 
than an hour. Under the GIM, the 
AHC generates higher values as high 
as 1.821 for a graph of 50 nodes in 
50 iterations. For a network of 
size 2000 nodes, the AHC generates 
1.71 clicks in approximately one 
hour. However for a graph of 4,500 
nodes notably under NIM, the 
AHC proves to be unfeasible taking 
5 hours to generate 1.250 clicks in 10 iterations. 
Hence we consider 
the AHC method as a moderately 
efficient method for obtaining near 
optimal solutions to the IM-RO 
problem. Taking into consideration 
(1) 
increasing the number of iterations 
increases the optimal expected 
number of 
clicks and running times and (2)   
utilizing ideal influence model parameters can 
generate higher optimal expected click 
values. \newline\indent
Table $(\ref{psocurrent})$ 
provides and analysis for the MPSO 
method on the IM-RO problem with 
both $c_{1}r_{1}$ and $c_{2}r_{2}$ 
set to 0.5. In
particular, we note the effect of 
increasing 
the swarm size, $\abs{n_{s}}$ and 
the
number of iterations $n$ on the 
optimal expected number of clicks.  For a network of 
50 users with $\abs{n_{s}} = 100$, 
and less than 10 iterations, the 
MPSO 
method generates ``good" results in 
minutes under the NIM.
However, for larger graphs, (i.e 
greater than 500), the MPSO 
converges slowly taking hours to 
converge to less accurate 
solutions. This is primarily  due to the fact that 
its running times increases 
significantly with its swarm size 
and number of iterations. From the 
analysis in 
Table(\ref{psocurrent}), we can 
conclude that the 
MPSO method is a fairly reasonable 
algorithm in terms of
achieving near-optimal solutions, 
however its running time is too 
slow making it unfeasible for 
large graphs. Moreover, it is unreliable in terms of accuracy since its state 
space consists of 
a set of optimal expected click 
values for predetermined users in
impression-to-stage 
allocations.\newline\indent

\begin{figure}[H]
 
\centering
  \includegraphics[width=90mm]
  {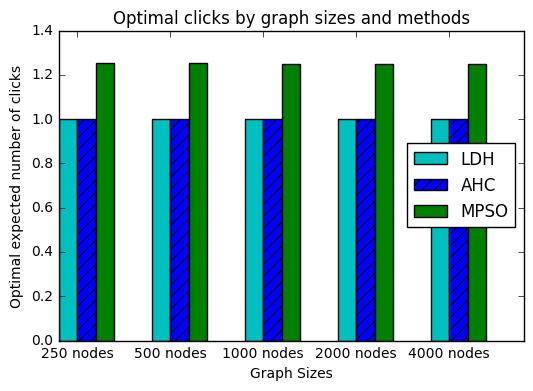}
  
 \caption{ NIM with 5 impressions in 2 stages, $\alpha=0.25$}
 \label{comparisonopt}

\end{figure}

\begin{figure}[H]
\centering
  \includegraphics[width=90mm]{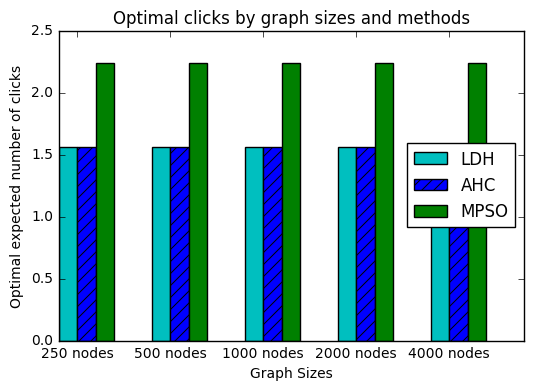}
\caption{GIM with 5 impressions in 2 stages, $\alpha=10$ }
\label{comparisonopt2}

\end{figure}

We observe Figure 
(\ref{comparisonopt}) and 
Figure $(\ref{comparisonopt2})$ 
and 
note the effect of varying 
$\alpha$ on all three methods.
The results indicate that the LDH 
and AHC generate identical optimal 
expected number of click values on 
various
synthetic networks. We consider 
Figure 
$(\ref{comparisonopt2})$
when $\alpha=10$ and highlight the 
significant increase in the 
optimal expected number of clicks 
from 1.0 to 1.5. These results 
have considerable gains for any 
OSN 
advertiser and significant 
implications for 
the choice of influence models and 
 the effect of optimizing influence model 
parameters in  
maximizing the expected number of 
clicks. Another reason for the 
similarity in  performance of the LDH and AHC algorithms can be attributed to the similarity in the synthetic networks each being generated by the same random number generator. We 
 note that the
MPSO algorithm  generates 
the highest expected number of 
clicks 
for all graph sizes however  its running time is too slow for large graphs, this result is further supported in our scalability analysis.

\subsection{Scalability}

To evaluate the scalability, the 
sizes of synthetic networks were 
doubled from 
250, 500, 1000,..., up to 4000 
nodes.

\begin{figure}[H]
 \begin{minipage}[b]{0.6\textwidth}
\centering
 \includegraphics[width=\textwidth]
  {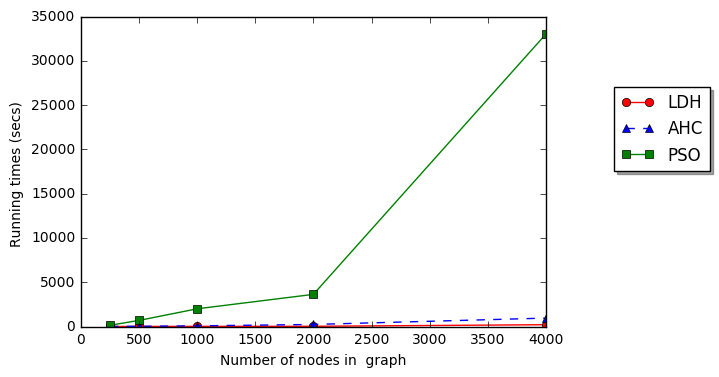}
 \caption{Regular scale}
 \label{scalability}
 \end{minipage}
\hfill
 \begin{minipage}[b]{0.6\textwidth}
\centering
\includegraphics[width=\textwidth]{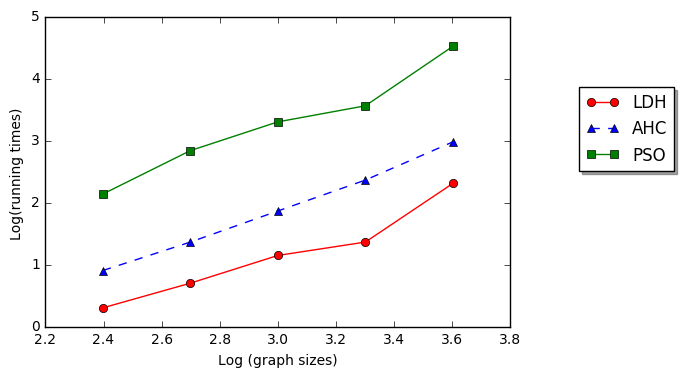}
 \caption{Log-Log Scale}
 \label{scalabilitylog}
 \end{minipage}
\end{figure}

Figures 
(\ref{scalability}) and 
(\ref{scalabilitylog}) 
demonstrate the results of the 
running 
times of the 
LDH, AHC and MPSO methods on a 
regular scale and log-log scale 
respectively. From 
the result 
in Fig 
(\ref{scalability}), we can 
clearly 
deduce that the 
PSO algorithm is not scalable 
since its 
running times 
is in the hour range for 2000 
nodes 
making 
it unfeasible to run on larger 
graphs. We 
also consider the high degree of 
the 
graphs generated  
by the psuedo random 
number 
generator allowing them to be 
suitable 
indicators for relatively any 
dataset. Figure 
(\ref{scalabilitylog}) provides 
a  further differentiation 
between 
the 
algorithms. From this results we 
conclude 
that all three algorithms have 
similar slopes, however the LDH and AHC
has both 
a good slope and intercept 
making them 
suitable for large graphs with 
at least 
thousands of nodes and edges.

\subsection{Performance on real-
world OSNs}

\begin{figure}[H]
\centering
  \includegraphics[width=100mm]{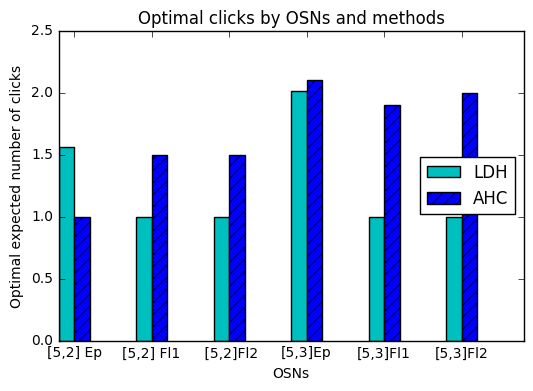}
 \caption{Real World Datasets}
 \label{realworldfig}

\end{figure}

We compare the computational time 
and optimal expected number of 
clicks generated by the LDH and 
AHC heuristics on two real-world 
OSNs under the GIM with model 
parameters $\alpha=5$, 
$p_{0}=0.25$ and 10 iterations. 
Table 
$(\ref{realworld})$ and Figure 
$(\ref{realworldfig})$
indicate that the performance of the 
LDH is considerably better or at 
least as good as the 
AHC heuristic in terms of the optimal 
number of clicks generated on the 
Epinions dataset whilst 
the AHC generates significantly 
higher optimal expected click values
on the Flickr dataset.  
We attribute these results to the 
design of the LDH being more suited 
to the structure of the OSN
Epinions and less to the structure of 
Flickr. Indeed, while both the 
LDH and the 
AHC 
heuristic achieve near optimal 
solutions in a run time of under 30 
minutes even for a network of 20,21 
users, the LDH attains the 
optimal values in seconds for all 
three networks.
In general, the AHC is 
well suited to both Epinions and 
Flickr  OSNs  in 
terms of its accuracy and running 
times. For a problem involving 5 
impressions, the optimal expected 
number of clicks generated is at 
least 2. 
Although the LDH generates similar 
results for a problem involving 5 
impressions 
on the Epinions dataset, the optimal 
expected number of clicks generated from 
the Flickr dataset is 1  even when there 
is an increase the number of stages.

\begin{table}[t]
\centering
\caption{Results on real-world OSN }
\label{realworld}       

\begin{tabular*}{\linewidth}{@{\extracolsep{\fill}}p{0.05\linewidth}p{0.05\linewidth}p{0.05\linewidth}p{0.05\linewidth}p{0.05\linewidth}p{0.1\linewidth}@{}}

\toprule
Method &OSN & Impressions & Stages  & 
Optimal Clicks & Time (secs)  \\

\midrule
LDH& Ep,4,382  &5 & 2 & 1.56 &4.4\\
 && 10 & 2& 3& 4.7\\
 & & 50 & 2& 13 & 4.7\\
 & &100 & 2& 25.5& 4.9\\
 & &200 & 2& 50.5& 5.5\\
 & &  5& 3 & 2.01& 30  \\
& Fl1,11,098 & 5 & 2 & 1& 12.6\\
& &10& 2 &2.25& 10\\
& &50 & 2 &12.25& 10 \\
&& 100 & 2 & 24.75 & 10  \\
&& 200 & 2&49.75& 10. \\
& &5& 3 & 1& 85\\
& Fl2,20,217 &5  & 2 & 1&15 \\
& &10& 2 &2.25& 15\\
& &50 & 2 &12.25& 16 \\
&& 100 & 2 & 24.75 & 15  \\
&& 200 & 2&49.75& 16 \\
& &5& 3 & 1& 83\\
AHC & Ep,4,382 &5 &2  &1  &20 \\
&&10 & 2&3.3&33  \\
&&50 & 2& 13& 389\\
&& 100 & 2 & 25.5 & 40\\
& &200 & 2& 50.6&42 \\
& & 5 & 3 & 2.1 & 45\\
& Fl1,11,098 & 5 & 2  & 1.5& 56\\
& &10& 2 &2.8 & 65\\
& &50 & 2 &12.9 & 65 \\
&& 100 & 2 & 25.3 &  71  \\
&& 200 & 2& 50.3 & 97 \\
& &5& 3 & 1.9 & 603\\
& Fl2,20,217 & 5 & 2 & 1.55& 163\\
& &10& 2 &3.25& 1701 \\
& &50 & 2 &13.13& 168 \\
&& 100 & 2 & 26.2 & 160  \\
&& 200 & 2&51.6& 157 \\
& &5& 3 &2& 1,298\\

\bottomrule
\end{tabular*}

\end{table}

The LDH 
and AHC heuristic exhibit good 
performance and are orders of 
magnitude 
faster than the SDP method.
The results for these heuristics 
suggest that advertising companies 
can target the optimal users to 
market (or spread information) to in 
OSNs in a way that can generate 
predictable and lucrative gains  
for socio-economic advancement.

\section {Conclusion}\label{sec:6}

We provide a novel
approach to influence 
maximization which until now has been 
primarily used in resource 
allocation and shortest path 
problems. We divert 
from previous approaches to 
influence maximization based on 
the theory of submodular 
functions and adopt a novel and 
practical 
decision-making approach geared 
towards 
maximizing clicks an revenue 
among users of an OSN. 
Hence we redefine the 
problem as IM-RO and introduce 
SDP as the method in which this 
problem can be solved. We first 
reviewed the properties of the  
SDP method on small synthetic 
networks and highlight the lucrative 
advantages that our method poses 
to advertising companies in terms 
of generating revenue and 
optimizing clicks. Due to the 
complexity of the SDP method, we 
sought to obtain heuristics which 
achieved near optimal solutions 
in considerably less 
time.\newline\indent
We second, proposed three 
heuristics, the LDH, AHC and MPSO 
algorithms which exploited the 
multistage 
attribute of the SDP method 
whilst reducing its complexity. In 
addition to achieving 
near-optimal solutions, all 
three methods were found to be 
orders of magnitude faster than 
the SDP method.  We provided a 
scalability analysis and
evaluated our proposed heuristics 
on synthetic networks of various 
sizes and two real-world OSNs, 
Flickr and Epinions. The LDH and 
AHC are shown to be well-suited 
heuristics for the SDP method in 
terms of 
their  accuracy, scalability and 
running times. 
The AHC is a more efficient 
heuristic than the  LDH since it 
outperforms the LDH in terms of 
accuracy and running times 
for the two real-world OSNs.  \newline\indent
We confirmed that the GIM 
exceeded the NIM in generating 
optimal expected 
number of click with 
approximately 
the same computational times.  
 It was shown that 
increasing $\alpha$ within both 
influence models significantly 
increased the optimal expected 
number of clicks even when 
$p_{0}$ remained small eg. 
$p_{0}=0.25$. This result 
provides substantial implications 
for the potential gains in 
obtaining ideal influence models 
and optimizing their associated 
model 
parameters. \newline\noindent
Our 
immediate future work is to 
provide an extensive analysis  on 
our influence models  and how 
their parameters affect the IM-RO 
problem. It is also necessary to 
obtain accurate estimates of the 
influence model parameters 
through statistical and data 
mining techniques in order to 
improve on the optimality of the 
expected number of 
clicks. \newline\indent
As a immediate consequence of  
approaching the IM 
problem through a decision-making 
perspective there are multiple 
directions for future work, both 
in terms of optimization 
(approximate dynamic programming methods) and 
data science. The 
results presented provide an 
evaluation for our methods on 
large networks. It is also 
necessary to derive an upper 
bound on the objective function 
in order to determine how well  
our methods perform on these
large networks. Another direction 
for future work
related to influence maximization 
is 
to obtain the influence spread 
for the IM-RO problem where the 
influence spread is defined as a function on 
the number of stages of the 
problem. Our future work also includes exploring this
applications in  fields 
of healthcare, 
communication, epidemiology, 
education, and agriculture.

\end{document}